\documentclass[12pt, twoside, leqno]{article}

\usepackage{amsmath,amsthm}
\usepackage{amssymb}
\usepackage{bbm}

\usepackage{enumitem}

\newtheorem{Theorem}{Theorem}[section]
\newtheorem{Corollary}[Theorem]{Corollary}
\newtheorem{Lemma}[Theorem]{Lemma}

\newtheorem{Question}[Theorem]{Question}

\theoremstyle{Definition}
\newtheorem{Definition}[Theorem]{Definition}
\newtheorem{Remark}[Theorem]{Remark}

\numberwithin{equation}{section}

\newcommand{\R}{\mathbb{R}}

\newcommand{\sgn}{{\rm sgn}}

\newcommand{\E}{\mathbb{E}}

\newcommand{\eps}{\varepsilon}

\def\IND{\mathbbm{1}}

\begin{document}


\baselineskip=17pt


\title{Extending the scope of the small-ball method}

\author{Shahar Mendelson \\
Mathematical Science Institute, \\
The Australian National University, \\
Canberra, Australia \\
and  \\
LPSM, Sorbonne University,
\\
Paris, France \\
E-mail: shahar.mendelson@anu.edu.au
}

\date{}

\maketitle


\renewcommand{\thefootnote}{}

\renewcommand{\thefootnote}{\arabic{footnote}}
\setcounter{footnote}{0}


\begin{abstract}
The small-ball method was introduced as a way of obtaining a high probability,
isomorphic lower bound on the quadratic empirical process, under weak assumptions on
the indexing class. The key assumption was that class members satisfy a uniform
small-ball estimate: that $Pr(|f| \geq \kappa\|f\|_{L_2}) \geq \delta$ for given
constants $\kappa$ and $\delta$.

Here we extend the small-ball method and obtain a high probability, almost-isometric (rather than isomorphic) lower bound on the quadratic empirical process. The scope of the result is considerably wider than the small-ball method: there is no need for class members to satisfy a uniform small-ball condition, and moreover, motivated by the notion of tournament learning procedures, the result is stable under a `majority vote'.
\end{abstract}

\section{Introduction}
In this article we study a more general version of the following question:
\begin{Question} \label{qu:main}
Let $F$ be a class of functions defined on a probability space $(\Omega,\mu)$, let $X$
be distributed according to $\mu$ and consider a sample $X_1,...,X_N$, consisting of
$N$ independent copies of $X$. Find a high probability, lower bound on
$\theta=\theta(r)$, defined by
\begin{equation} \label{eq:standard-lower-1}
\theta=\inf_{\{f \in F : \|f\|_{L_2} \geq r\}} \frac{1}{N}\sum_{i=1}^N
\frac{f^2(X_i)}{\|f\|_{L_2}^2},
\end{equation}
for a value of $r$ that is as small as possible.
\end{Question}
The obvious implication of \eqref{eq:standard-lower-1} is if $f \in F$ and
$\|f\|_{L_2} \geq r$ then
$$
\frac{1}{N}\sum_{i=1}^N f^2(X_i) \geq \theta \|f\|_{L_2}^2,
$$
which is an `isomorphic' lower bound on the quadratic empirical process.

Lower bounds on \eqref{eq:standard-lower-1} play an important role in applications in
probability (e.g., the smallest singular value of a random matrix with iid rows),
geometry (for example, estimates on the Gelfand widths of convex bodies
\cite{MR941809,MR845980,MR3331351,MR2373017}), and statistics.

\vskip0.3cm

The standard way of estimating \eqref{eq:standard-lower-1} is by two-sided
concentration, that is, by obtaining a high probability \emph{upper bound} on
\begin{equation} \label{eq:two-sided-intro}
\sup_{\{f \in F : \|f\|_{L_2} \geq r\}} \left|\frac{1}{N}\sum_{i=1}^N
\frac{f^2(X_i)}{\|f\|_{L_2}^2}-1\right|.
\end{equation}
Estimates of this type are called  \emph{ratio-limit theorems} (see
\cite{MR2073436,MR2243881} and references therein). However, a nontrivial ratio-limit
theorem is possible only if class members have well-behaved tails, and even then
obtaining the two-sided estimate is rather involved (see, e.g.,
\cite{MR2373017,Men-multi16}).

The fact that a high probability, two-sided estimate as in \eqref{eq:two-sided-intro}
is false without assuming that class members have well-behaved tails can be seen by
considering what happens for a single function: given a square-integrable function
$f$, the probability that
$$
\frac{1}{N}\sum_{i=1}^N f^2(X_i) \leq C \|f\|_{L_2}^2
$$
may be small; specifically, it need not be better than the outcome of Chebychev's inequality. Even
if one allows for large values of $C$ the situation remains the same: for example, it
is straightforward to construct a function $f$ on the unit sphere of $L_2(\mu)$ such
that
$$
Pr\left(\frac{1}{N} \sum_{i=1}^N f^2(X_i) \geq N\right) \geq \frac{c_1}{N}.
$$

In contrast, a lower bound of the form
\begin{equation} \label{eq:intro-lower}
\frac{1}{N}\sum_{i=1}^N f^2(X_i) \geq c \|f\|_{L_2}^2
\end{equation}
is almost universal and holds with very high probability under minimal assumptions on
$f$:
\begin{Definition} \label{def:small-ball-condition}
The function $f$ satisfies a \emph{small-ball condition} with constants $\kappa>0$ and
$0<\delta<1$ if
$$
Pr(|f| \geq \kappa \|f\|_{L_2}) \geq \delta.
$$
\end{Definition}
All that a small-ball condition implies is that $f$ does not assign too much weight to
a small neighbourhood of $0$; it does not mean that $f$ has a well behaved tail, and
in particular, it does not exclude the possibility that $f$ does not have any moment
beyond the second one. As it happens, a small-ball condition is enough to ensure that
the lower bound \eqref{eq:intro-lower} holds with very high probability for a
well-chosen constant $c$. Indeed, a standard binomial estimate shows that with
probability at least $1-2\exp(-c_1\delta N)$,
$$
|\{i : |f(X_i)| \geq \kappa \|f\|_{L_2} \}| \geq \frac{\delta N}{2};
$$
therefore, on that event,
$$
\frac{1}{N} \sum_{i=1}^N f^2(X_i) \geq \frac{\delta}{2} \kappa^2  \|f\|_{L_2}^2.
$$

\vskip0.4cm

This overwhelming difference between the upper and lower bounds on
$\frac{1}{N}\sum_{i=1}^N f^2(X_i)$ motivated the introduction of the \emph{small-ball
method} \cite{MR3364699,MenACM}. It has led to a lower bound on
\eqref{eq:standard-lower-1} under the assumption that the class in question satisfies a small-ball property---that there are constants $\kappa>0$ and $0<\delta <1$ such that
for every $f \in F$, $Pr(|f| \geq \kappa \|f\|_{L_2}) \geq \delta$. To formulate this lower bound let
$$
{\rm star}(H,f) = \{\lambda h + (1-\lambda) f : h \in H, \ 0 \leq \lambda \leq 1\}
$$
which is the star-shaped hull of $H$ with $f$. Also, from here on denote by $(\eps_i)_{i=1}^N$ independent, symmetric
$\{-1,1\}$-valued random variables that are also independent of $(X_i)_{i=1}^N$; $D$
is the unit ball in $L_2(\mu)$; and $S$ is the corresponding unit sphere.

\begin{Theorem} \label{thm:SBM-intro} \cite{MenACM}
There exist absolute constants $c_1$ and $c_2$ for which the following holds.
Let $H \subset L_2(\mu)$ and assume that for every $h \in H$, $Pr(|h| \geq \kappa
\|h\|_{L_2}) \geq \delta$. If $r>0$ satisfies that
$$
\E \sup_{h \in {\rm star}(H,0) \cap rS} \left|\frac{1}{\sqrt{N}}\sum_{i=1}^N \eps_i
h(X_i) \right| \leq c_1 \kappa \delta r \sqrt{N},
$$
then with probability at least $1-2\exp(-c_2 \delta N)$
$$
\inf_{\{h \in H \ : \ \|h\|_{L_2} \geq r\}} \left|\left\{i: |h(X_i)| \geq
\frac{\kappa}{2} \|h\|_{L_2} \right\}\right| \geq \frac{N \delta}{4}.
$$
In particular, on the same event,
$$
\inf_{\{h \in H \ : \ \|h\|_{L_2} \geq r\}} \frac{1}{N} \sum_{i=1}^N
\frac{h^2(X_i)}{\|h\|_{L_2}^2} \geq \frac{\kappa^2 \delta}{16}.
$$
\end{Theorem}

Theorem \ref{thm:SBM-intro} has many applications, but our focus here is on three
directions in which it is natural to extend its scope:
\begin{description}
\item{$\bullet$} Although a small-ball property is a rather minimal condition on a
    class, there are still important situations in which it is not satisfied. For
    example, if the class $H$ is a bounded subset of $L_p$ for some $p>2$, it need
    not satisfy a small-ball property. In fact, even if $p=\infty$ and class members
    are bounded almost surely by $1$, the best possible choice of $\delta$ for a
    function $h$ may be as bad as $\|h\|_{L_2}^2$. As an example, fix $0 < \rho <1$ and let $h$ be a $\{0,1\}$-valued such that $Pr(h=1)=\rho$. Therefore, $\|h\|_{L_2}^2 = \rho$ and $Pr(|h| \geq \kappa \|h\|_{L_2}) = \|h\|_{L_2}^2$ for any $0<\kappa \leq 1/\sqrt{\rho}$, and $0$ otherwise.

    With that in mind, one would like to find a version of Theorem
    \ref{thm:SBM-intro} that is strong enough to deal with more general situations
    than classes that satisfy a small-ball property.

\item{$\bullet$} Results that are based on a small-ball property are of an
    isomorphic nature. The best that one can hope for is that if $h \in H$ and
    $\|h\|_{L_2} \geq r$, then
    $$
    \frac{1}{N} \sum_{i=1}^N h^2(X_i) \geq c(\kappa,\delta) \|h\|_{L_2}^2,
    $$
    where $c(\kappa,\delta)$ depends only on the small-ball parameters $\kappa$ and
    $\delta$. The fact that $c$ is not close to $1$ is unfortunate but unavoidable.
    On the other hand, at times one requires an almost isometric lower bound, with
    $c=1-\xi$ for a small $\xi$. Therefore, the second extension of the small-ball
    method is to ensure that for a fixed $0<\xi<1$ which can be (almost) arbitrarily
    small, and with high probability one has
    $$
    \inf_{\{h \in H : \|h\|_{L_2} \geq r\}} \frac{1}{N}\sum_{i=1}^N
    \frac{h^2(X_i)}{\|h\|_{L_2}^2} \geq 1-\xi.
    $$
\item{$\bullet$} The final extension is motivated by \emph{tournaments}
    \cite{LugMen16,Men17}. Tournaments are statistical procedures that attain the
    optimal accuracy/confidence tradeoff for (almost) any prediction problem
    relative to the squared loss. Roughly and somewhat inaccurately put, consider a class of functions $F$ and an unknown random variable $Y$. One would like to estimate $Y$ by some $f \in F$ whose $L_2$ distance to $Y$ is almost the best possible in $F$. The data one is given to perform this task is an iid sample $(X_i,Y_i)_{i=1}^N$, selected according to the joint distribution of $X$ and $Y$.

    To identify which of the class members is almost optimal, one splits the given sample $(X_i,Y_i)_{i=1}^N$ to $n$ coordinate blocks $(I_j)_{j=1}^n$, each one of cardinality $m$; and for any $f,h \in F$ one compares the $n$ empirical errors
    $$
    \frac{1}{m}\sum_{i \in I_j} (f(X_i)-Y_i)^2 \ \ \ {\rm and} \ \ \
    \frac{1}{m}\sum_{i \in I_j} (h(X_i)-Y_i)^2.
    $$
 Based on the outcomes, one nominates the winner in this ``statistical match" between $f$ and $h$, and the key feature of this ``tournament" is that if $f$ ``wins" in this comparison then $\E(f(X)-Y)^2 < \E (h(X)-Y)^2$. The procedure selects a function that wins all of its
    matches\footnote{In actual fact, the choice of a winner of a tournament is more
    involved. The reason is that when the functions are too close to each other, the
    outcome of the statistical match between them is unreliable. As a result, the ``winner" of the tournament need not be the actual minimizer, but rather a function that is almost as good. For more details, see \cite{LugMen16,Men17}.}. As it happens, at
    the heart of the analysis of tournament procedures is the following question:
\end{description}
\begin{Question} \label{Qu:modified-main}
Let $H \subset L_2(\mu)$. Fix an integer $n$ and set $(I_j)_{j=1}^n$ to be the
decomposition of $\{1,...,N\}$ to $n$ blocks of equal size which is denoted by $m$.
Given $0<\xi<1$ and $0<\eta<1$, find $r>0$ that is as small as possible such that with
high probability, for any $h \in H$ with $\|h\|_{L_2} \geq r$
$$
\left|\left\{ j : \frac{1}{m} \sum_{i \in I_j} h^2(X_i) \geq (1-\xi)\|h\|_{L_2}^2
\right\} \right| \geq (1-\eta)n.
$$
\end{Question}

Question \ref{Qu:modified-main} is significantly harder than Question \ref{qu:main}:
for every function in the class whose $L_2$ norm is not too small one must show an
almost isometric lower bound that holds for a large majority of the coordinate blocks
$I_j$. Clearly, when $n=1$ and $\eta=0.5$ (or any other constant smaller than $1$) and
$\xi$ is a constant that need not be small, Question \ref{Qu:modified-main} reverts to
Question \ref{qu:main}.

\vskip0.4cm

Here we answer Question \ref{Qu:modified-main} without assuming that the class
satisfies a small-ball property, thus extending Theorem \ref{thm:SBM-intro} in all the
three directions we outlined. The estimate holds, for example, for bounded
subsets of $L_p$; for classes that satisfies an $L_q-L_2$ norm equivalence for some
$q>2$; and when the class satisfies a uniform integrability condition as in \cite{Men17}.

\vskip0.4cm

We end this introduction with some notation. Throughout the article, absolute
constants are denoted by $c,c_1,...$ and $C,C_1,....$. Their value may change from
line to line. $c_p$ or $c(p)$ means that the constants depend only on the parameter
$p$. We write $a \sim b$ when there are absolute constants $c$ and $C$ such that $ca
\leq b \leq Ca$, and $a \lesssim b$ if only a one-sided inequality holds; $a \sim_p b$
and $a \lesssim_p b$ implies that the constants depend only on the parameter $p$.

\section{Beyond the small-ball condition}
Before one can extend the small-ball method one must first identify a notion that can
replace the small-ball condition. To that end, let us examine the way in which a
small-ball condition is used to establish the wanted lower bound.

Given $X_1,...,X_m$, a small-ball condition with constants $\kappa$ and $\delta$
implies that with very high probability ($1-2\exp(-c\delta m)$), there are at least
$\delta m /2$ indices $i$ such that $|h(X_i)| \geq \kappa \|h\|_{L_2}$. Thus, not
only is
\begin{equation} \label{eq:sbm-implication}
\frac{1}{m}\sum_{i=1}^m h^2(X_i) \gtrsim_{\kappa,\delta} \|h\|_{L_2}^2,
\end{equation}
but \eqref{eq:sbm-implication} is stable: discarding a small proportion of the
coordinates $\{1,...,m\}$ does not ruin the lower bound.

The notion used in what follows captures these features: not only is
$\frac{1}{m}\sum_{i=1}^m h^2(X_i)$ large enough with high probability, it remains
large if any subset of $\{1,...,m\}$ of a reasonable cardinality is discarded from the sum.

\begin{Definition} \label{def:stable-lower}
A function $h$ satisfies a stable lower bound with parameters $(\xi,\ell,k)$ for a
sample of cardinality $m$  if with probability at least $1-2\exp(-k)$, for any $J
\subset \{1,...,m\}$, $|J| \leq \ell$ one has
$$
\frac{1}{m}\sum_{i \in J^c} h^2(X_i) \geq (1-\xi)\|h\|_{L_2}^2.
$$
In what follows we do not specify the cardinality of the coordinate block in question
(it is denoted by $m$ throughout the article); instead we just say that $h$ satisfies
a stable lower bound with parameters $(\xi,\ell,k)$.
\end{Definition}

\subsubsection*{Stability and geometry}
The notion of a stable lower bound has a geometric interpretation. The fact that
$$
\frac{1}{m} \sum_{i=1}^m h^2(X_i) \geq (1-\xi) \|h\|_{L_2}^2
$$
obviously means that the random vector $v = (h(X_i))_{i=1}^m$ is located outside the
Euclidean ball $(1-\xi)^{1/2}\sqrt{m} \|h\|_{L_2} B_2^m$. Also, with constant
probability and in expectation, $\|v\|_2 \lesssim \sqrt{m}\|h\|_{L_2}$, placing $v$
inside a ``shell" of inner and outer radius $\sim \sqrt{m}\|h\|_{L_2}$. However, all that information says very little about the coordinate
distribution of the vector: the fact that $v$ has a Euclidean norm of order $\sqrt{m}\|h\|_{L_2}$ does not rule out the possibility that all of its `mass' is
concentrated at a single coordinate. In contrast, a stable lower bound implies that
the vector $v$ is well-spread: its $m-\ell$ smallest coordinates still carry
significant mass. This fact has a probabilistic implication as well: the (conditional) Bernoulli random variable
$\sum_{i=1}^m \eps_i h(X_i)=\sum_{i=1}^m \eps_i v_i$ exhibits a gaussian-like
behaviour. Indeed,  it is well known (see \cite{MR1244666}) that for every $p \geq 2$
and every $x \in \R^m$,
$$
\|\sum_{i=1}^m \eps_i x_i\|_{L_p}
\sim \sum_{i \leq p} |x_i^*| + \sqrt{p}
\left(\sum_{i > p} (x_i^*)^2 \right)^{1/2},
$$
where $(x_i^*)_{i=1}^m$ denotes the nonincreasing rearrangement of $(|x_i|)_{i=1}^m$.
If all the mass of $x$ is concentrated at a single coordinate then $\|\sum_{i=1}^m
\eps_i x_i \|_{L_p} \sim \|x\|_2$, whereas for a gaussian like behaviour one would
expect to have that $\|\sum_{i=1}^m \eps_i x_i\|_{L_p}$ is equivalent to $\sqrt{p}\|x\|_2$. Thanks to this notion of
stability it follows that if $v=(h(X_i))_{i=1}^m$ then with probability at least
$1-2\exp(-k)$,
$$
\sum_{i \geq \ell} (v_i^*)^2 \geq (1-\xi)m\|h\|_{L_2}^2,
$$
and on that event,
\begin{equation} \label{eq:slb-implies-Bernoulli}
\|\sum_{i=1}^m \eps_i v_i \|_{L_p} \gtrsim \sqrt{p} \sqrt{m} \|h\|_{L_2} \sim
\sqrt{p}\E\|v\|_2 \ \ \ {\rm for} \ \ \ 2 \leq p \leq \ell.
\end{equation}

\begin{Remark}
It should be stressed that Definition \ref{def:stable-lower} is very different from concentration. If the smaller coordinates of the nonincreasing rearrangement $(v_i^*)_{i \geq \ell}$ of a typical realization $v=(h(X_i))_{i=1}^m$ have `enough
mass' then $h$ satisfies a stable lower bound. However, the larger coordinates
$(v_i^*)_{i=1}^\ell$ can completely destroy any hope of a reasonable upper estimate on
$\frac{1}{m}\sum_{i=1}^m h^2(X_i)$, making two-sided concentration impossible.
\end{Remark}

\subsection{Examples of a stable lower bound}

To put the notion of a stable lower bound in some context, let us show that there are
many natural situations in which it holds.

\subsubsection*{A bounded function}
\noindent Let $h$ be a function that is bounded almost surely by $M$. As the next
lemma shows, $h$ satisfies a stable lower bound.
\begin{Lemma} \label{lemma:slb-bounded}
There are absolute constants $c_0$ and $c_1$ for which the following holds. Let $h$ be
a function that is bounded almost surely by $M$. For any $0<\xi<1$, $h$ satisfies a
stable lower bound with parameters $(\xi,\ell,k)$ for
$$
\ell = c_0 m\xi \frac{\E h^2}{M^2} \ \ \ {\rm and} \ \ \ k = c_1 m \xi^2 \frac{\E
h^2}{M^2}.
$$
\end{Lemma}

\proof
Applying Bernstein's inequality, it follows that
$$
Pr \left( \left|\frac{1}{m} \sum_{i=1}^m h^2(X_i) - \E h^2 \right| > u \right) \leq
2\exp\left(-cm \min\left\{\frac{u^2}{\E h^4}, \frac{u}{\|h^2\|_{L_\infty}}
\right\}\right).
$$
Note that $\E h^4 \leq M^2 \E h^2$ and $\|h^2\|_{L_\infty} \leq M^2$. Setting $u =
(\xi/2) \E h^2$ it is evident that with probability at least $1-2\exp(-c_1 m \xi^2 \E
h^2/M^2)$,
$$
\frac{1}{m}\sum_{i=1}^m h^2(x_i) \geq \left(1-\frac{\xi}{2}\right) \E h^2.
$$
The contribution to the sum of the $\ell$ largest coordinates is at most $\ell M^2/m$,
which is at most $(\xi/2) \E h^2$ provided that $\ell \leq m \xi \E h^2/2M^2$, as
claimed.
\endproof

Lemma \ref{lemma:slb-bounded} is not very surprising because empirical means of a bounded
function exhibit a two-sided concentration around the true mean, which in return
implies a stable lower bound. Still, this example is of interest because a bounded function need not satisfy a
nontrivial small-ball property.

\vskip0.3cm

When leaving the bounded realm the situation is not as straightforward. And the other
examples presented here are of that nature: situations in which a stable lower bound holds but there
is no hope for a two-sided concentration of the empirical mean.

\subsubsection*{Tail cutoff}
Because our interest lies in obtaining a lower bound, truncating the function is a
possible approach. And, there is a natural location in which the function should be
truncated:
\begin{Definition} \label{def:unifrom-integrability-single}
For a function $h$ and $0<\xi<1$, set
$$
M(h,\xi) = \inf \left\{ t: \E h^2 \IND_{\{|h| > t \} } \leq \frac{\xi}{2} \E h^2
\right\}.
$$
\end{Definition}
In other words, $M(h,\xi)$ is the smallest level at which the
truncated function $w=h \IND_{\{|h| \leq t \}}$ still has a significant $L_2$ norm:
$\E w^2 \geq (1-\xi/2)\E h^2$. Applying Lemma \ref{lemma:slb-bounded} to the
truncated function $h \IND_{\{|h| \leq M(h,\xi) \}}$, one has the following:

\begin{Corollary} \label{cor:slb:integrability}
There are absolute constants $c_0$ and $c_1$ for which the following holds. If $0<\xi
<1$ and $M=M(h,\xi)$ then $h$ satisfies a stable lower bound with parameters
$(\xi,\ell,k)$ for
$$
\ell=c_0 m \xi  \frac{\E h^2}{M^2} \ \ \ {\rm and} \ \ \ k = c_1 m \xi^2  \frac{\E
h^2}{M^2}.
$$
\end{Corollary}

An important example of a tail cutoff, which has been studied in \cite{Men17} in the
context of tournaments, is when one is given a class of functions $H$ such that for any
$h \in H$, $M(h,\xi) \leq \kappa(\xi)\|h\|_{L_2}$.
\begin{Definition} \label{def:uniform-integrablity}
A class $H$ satisfies a uniform integrability condition if for every $0<\xi<1$ there
is $\kappa(\xi)$ such that for every $h \in H$,
$$
\E h^2 \IND_{\{|h| \geq \kappa(\xi) \|h\|_{L_2}\}} \leq \frac{\xi}{2} \|h\|_{L_2}^2.
$$
\end{Definition}
Again, it is standard to verify that each $h \in H$ satisfies a stable lower bound with constants
$$
\ell \sim m\frac{\xi}{\kappa^2(\xi)} \ \ \ {\rm and} \ \ \ k \sim m
\frac{\xi^2}{\kappa^2(\xi)}.
$$
\vskip0.3cm
Once one has more information on $h$, an improved
estimate on the cutoff point $M(h,\xi)$ is possible, which also affects the
way $\sum_{i=1}^m h^2 \IND_{\{|h| \leq M\}} (X_i)$ concentrates around its mean. Two
such examples are when $\|h\|_{L_p} \leq L$ and when there is norm equivalence between
the $L_q$ and $L_2$ norms, i.e., when $\|h\|_{L_q} \leq L \|h\|_{L_2}$.

\subsubsection*{A function bounded in $L_p$}
Let $h \in L_p$ for some $p>2$. To identify its cutoff point, let $q=p/2$ and set
$q^\prime$ to be the conjugate index of $q$. Then
\begin{equation*}
\E h^2 \IND_{\{|h| > t\}} \leq (\E h^{2q})^{1/q} (Pr(|h| > t))^{1/q^\prime} \leq (\E
|h|^p)^{1/q} \cdot \frac{ (\E |h|^p)^{1/q^\prime}}{t^{p/q^\prime}} = \frac{\E
|h|^p}{t^{p-2}}.
\end{equation*}
Therefore,
$$
M(h,\xi) \leq \left(\frac{2\|h\|_{L_p}^p}{\xi \|h\|_{L_2}^2}\right)^{1/(p-2)} =
2^{1/(p-2)} \|h\|_{L_p} \cdot \left(\frac{\|h\|_{L_p}^2}{\xi
\|h\|_{L_2}^2}\right)^{1/(p-2)},
$$
and one has
$$
\ell = c_0 m \xi  \frac{\E h^2}{M^2} \geq c_1(p) m\left(\frac{\xi
\|h\|_{L_2}^2}{\|h\|_{L_p}^2}\right)^{p/(p-2)}.
$$
To identify $k$, set $Z=h^2 \IND_{\{|h| \leq M\}}(X)$, and observe that
\begin{equation*}
\E Z^2 \leq
\begin{cases}
c_2(p) \|h\|_{L_p}^4 \left(\frac{\|h\|_{L_p}^2}{\xi
\|h\|_{L_2}^2}\right)^{(4-p)/(p-2)} &\mbox{if } 2 < p < 4,
\\
\|h\|_{L_4}^4  & \mbox{if } p \geq 4.
\end{cases}
\end{equation*}
Let $Z_1,...,Z_m$ be independent copies of $Z$. Applying Bernstein's inequality it
follows that
$$
\left| \frac{1}{m}\sum_{i=1}^m Z_i - \E Z \right| > \frac{\xi}{2} \|h\|_{L_2}^2
$$
with probability at most
\begin{equation*}
\begin{cases}
2\exp\left(-c_3(p)m\left(\frac{\xi
\|h\|^2_{L_2}}{\|h\|^2_{L_p}}\right)^{p/(p-2)}\right)   &\mbox{if } \ \ 2 < p < 4,
\\
2\exp \left(-c_4m \min\left\{\frac{\xi^2 \|h\|_{L_2}^4}{\|h\|_{L_4}^4},
\left(\frac{\xi \|h\|^2_{L_2}}{\|h\|^2_{L_p}}\right)^{p/(p-2)} \right\} \right) &
\mbox{if } \ \ p \geq 4.
\end{cases}
\end{equation*}
implying that one may set
\begin{equation*}
k =
\begin{cases}
c_3(p) m\left(\frac{\xi \|h\|^2_{L_2}}{\|h\|^2_{L_p}}\right)^{p/(p-2)}   &\mbox{if } \
\ 2 < p < 4,
\\
c_4 m \min\left\{ \left(\frac{\xi \|h\|_{L_2}^2}{\|h\|_{L_4}^2}\right)^2,
\left(\frac{\xi \|h\|^2_{L_2}}{\|h\|^2_{L_p}}\right)^{p/(p-2)} \right\}  & \mbox{if }
\ \ p \geq 4.
\end{cases}
\end{equation*}

\begin{Remark} \label{rem:k-large-p}
Note that if $p>4$ then $h^4 = h^\alpha h^{4-\alpha}$ for $\alpha=2(p-4)/(p-2)$. By
H\"{o}lder's inequality for $q=(p-2)/(p-4)$ and $q^\prime=(p-2)/2$, it follows that
$$
\E h^4 \leq (\E h^2)^{(p-4)/(p-2)} \cdot (\E |h|^p)^{2/(p-2)};
$$
therefore,
$$
\frac{\|h\|_{L_2}^4}{\|h\|_{L_4}^4} \geq
\left(\frac{\|h\|_{L_2}^2}{\|h\|_{L_p}^2}\right)^{p/(p-2)},
$$
and for $p \geq 4$ one may take
$$
k \sim m \xi^2 \left(\frac{\|h\|_{L_2}^2}{\|h\|_{L_p}^2}\right)^{p/(p-2)}.
$$
\end{Remark}

\subsubsection*{Norm equivalence}
Another useful example is when $h$ satisfies an $L_q-L_2$ norm equivalence, i.e, when
$\|h\|_{L_q} \leq L \|h\|_{L_2}$ for some constant $L$. It follows that
$$
M(h,\xi) \leq \left(\frac{2L^2}{\xi}\right)^{1/(q-2)} \|h\|_{L_q},
$$
and one may set
\begin{equation*}
\ell = c_1(q) m \left(\frac{\xi}{L^2}\right)^{q/(q-2)} \ \ \ {\rm and} \ \ \
k =
\begin{cases}
c_2(q) m \left(\frac{\xi}{L^2}\right)^{q/(q-2)}   &\mbox{if } \ \ 2 < q < 4,
\\
c_3  m \left(\frac{\xi}{L^2}\right)^2  & \mbox{if } \ \ q \geq 4.
\end{cases}
\end{equation*}

\section{The main result}
With the notion of a stable lower bound set in place and armed with the examples, let us formulate the main
result of this note. To that end, fix integers $m,n$ such that $N=mn$ and let $(I_j)_{j=1}^n$ be
the natural partition of $\{1,...,N\}$ to coordinate blocks of cardinality $m$. Recall
that $D$ is the unit ball in $L_2(\mu)$ and $S$ is the corresponding unit sphere. For
$F \subset L_2(\mu)$ denote by ${\cal M}(F,\rho D)$ the cardinality of a maximal
$\rho$-separated subset of $F$ with respect to the $L_2(\mu)$ norm.

\begin{Theorem} \label{thm:main-lower}
There exist absolute constants $c_0,c_1$ and $c_2$ for which the following holds.  Let
$H$ be star-shaped around $0$ (i.e., ${\rm star}(H,0)=H$) and for $r>0$ set $H_r = H
\cap rD$. Fix $0<\eta,\xi<1$ and let $r>0$ such that
\begin{description}
\item{$(1)$} Every $h \in H \cap rS$ satisfies a stable lower bound with parameters
    $(\xi/2,\ell,k)$, for $k \geq \max\{4,2\log(4/\eta)\}$.
\item{$(2)$} $\log{\cal M}(H \cap rS, c_0\sqrt{\eta} \xi r D) \leq \frac{\eta N}{16}
    \cdot \frac{k}{m}$.
\item{$(3)$} $\E \sup_{u \in (H_r-H_r) \cap c_0\sqrt{\eta} \xi r D}
    \left|\frac{1}{N} \sum_{i=1}^N \eps_i u(X_i) \right| \leq c_1\eta \xi r \cdot
    \sqrt{\frac{\ell}{m}}$.
\end{description}
Then with probability at least
$$
1-2\exp\left(-c_2 \eta N \min\left\{\frac{\ell}{m},\frac{k}{m}\right\}\right)
$$
we have
$$
\inf_{\{h \in H \ : \ \|h\|_{L_2} \geq r\}} \left| \left\{ j : \frac{1}{m}\sum_{i \in
I_j} h^2(X_i) \geq (1-\xi)\|h\|_{L_2}^2 \right\} \right| \geq (1-\eta)n.
$$
Moreover, the same assertion holds if one replaces Conditions $(2)$ and $(3)$ with
\begin{description}
\item{$(4)$} $\E \sup_{u \in H_r} \left|\frac{1}{N} \sum_{i=1}^N \eps_i u(X_i)
    \right| \leq c_3\eta \xi r
    \cdot\min\left\{\sqrt{\frac{\ell}{m}},\sqrt{\frac{k}{m}}\right\}$, where $c_3$
    is an absolute constant.
\item{(5)} If $h_1,h_2 \in H \cap rS$ and $\|h_1-h_2\|_{L_2} \geq c_0 \sqrt{\eta}
    \xi r$ then $h_1-h_2$ satisfies a stable lower bound with parameters
    $(1/2,\ell,k)$.
\end{description}
\end{Theorem}

\begin{Remark}
In what follows we only consider the more difficult case, in which $0<\xi \leq 1/2$,
and the required estimate is truly almost isometric rather than isomorphic. We omit
the proof of Theorem \ref{thm:main-lower} when $1-\xi$ is closer to $0$ (e.g., in the
situation explored in Theorem \ref{thm:SBM-intro} using the standard small-ball
method), which requires a minimal modification of the argument we do present.

Let us mention that Theorem \ref{thm:SBM-intro} is a straightforward outcome of Theorem \ref{thm:main-lower} (with slightly different constants) when $H$ satisfies a
small-ball property, making Theorem \ref{thm:main-lower} a true extension of the
small-ball method.
\end{Remark}

\vskip0.4cm
The sufficient condition described in the ``moreover" part of Theorem
\ref{thm:main-lower} can be far from optimal because Condition $(4)$ is significantly
more restrictive than the combination of Conditions $(2)$ and $(3)$, forcing one to
consider larger values of $r$. Indeed, standard examples of a stable lower bound
indicate that often $k \sim \xi \ell$. Therefore, taking the minimum between
$\sqrt{\ell/m}$ and $\sqrt{k/m}$ comes at a cost of $\sim \sqrt{\xi}$. Moreover, the
indexing set $H \cap rS$ may be much larger than $(H_r-H_r) \cap c\sqrt{\xi} r D$. Both
factors affect the outcome of Theorem \ref{thm:main-lower} when one is looking for a
sharp dependence on $\xi$ or when $\xi$ is very small---tending to $0$ with $N$.
However, when $\xi$ happens to be a fixed constant, the combination of Condition $(4)$
and Condition $(5)$ is a suitable replacement for Conditions $(2)$ and $(3)$.

\vskip0.3cm
It is straightforward to apply Theorem \ref{thm:main-lower} to any class of functions
whose members satisfy a stable lower bound. We chose to focus on one example: a class that is bounded in $L_p$ for some $2<p \leq \infty$.

%
%

\subsubsection*{Bounded subsets of $L_p$}

The proof of the following corollary is based on the ``moreover" part of Theorem \ref{thm:main-lower}:
\begin{Corollary} \label{cor:counded-functions}
Let $p>2$ and assume that $H$ is a bounded class in $L_p$, by $M_p$. Set $r>0$ such
that
$$
\E \sup_{h \in H \cap c_0 r D} \left|\frac{1}{N} \sum_{i=1}^N \eps_i u(X_i) \right|
\leq c_1r \left(\frac{r}{M_p}\right)^{p/(p-2)}.
$$
Then with probability at least
$$
1-2\exp\left(-c_2N \left(\frac{r^2}{M_p^2}\right)^{p/(p-2)}\right),
$$
one has
\begin{equation} \label{eq:assert-in-cor}
\inf_{\{h \in H \ : \ \|h\|_{L_2} \geq r\}} \left| \left\{ j : \frac{1}{m}\sum_{i \in
I_j} h^2(X_i) \geq (1-\xi)\|h\|_{L_2}^2 \right\} \right| \geq 0.99n;
\end{equation}
here
$c_0$ is a constant that depends only on $\xi$ and $c_1$ and $c_2$ depend only on
$\xi$ and $p$.

Moreover, for $p=\infty$, i.e., if every $h \in H$ satisfies that $\|h\|_{L_\infty}
\leq M$ and if
$$
\E \sup_{h \in H \cap c_0 r D} \left|\frac{1}{N} \sum_{i=1}^N \eps_i u(X_i) \right|
\leq c_1 \frac{r^2}{M},
$$
then with probability at least $1-2\exp(-c_2N (r^2/M^2))$, \eqref{eq:assert-in-cor}
holds; here
$c_0,c_1$ and $c_2$ depend only $\xi$.
\end{Corollary}

The case $p=\infty$ can be established using other methods that are based on Talagrand's concentration inequality for empirical processes indexed by bounded subsets in $L_\infty$ (see the formulation of Talagrand's theorem in what follows). However, for $p<\infty$ this concentration based argument is no longer valid and as a result estimates like \eqref{eq:assert-in-cor} where out of reach.

\section{Proof of Theorem \ref{thm:main-lower}}
The proof of Theorem \ref{thm:SBM-intro}, whose general path is followed here as well, is based on three components.
Firstly, an individual estimate that holds with high probability---specifically, that
with probability at least $1-2\exp(-c_0 m \delta)$,
$$
\frac{1}{m}\sum_{i=1}^m h^2(X_i) \geq c_1(\kappa,\delta) \|h\|_{L_2}^2;
$$
secondly, that this estimate is stable: discarding a reasonable number of coordinates
does not significantly affect the sum; and finally, a second type of stability: if
$f,h$ are close then the vector $\left((f-h)(X_i)\right)_{i=1}^N$ does not have many
large coordinates. Once these properties are established, the high probability
individual estimate leads to uniform control over a net, and the two notions of
stability allow one to pass from the net to the entire class.

The same ideas are used in the proof of Theorem \ref{thm:main-lower}. Because the
claim is homogeneous and $H$ is star-shaped around $0$, it suffices to prove Theorem
\ref{thm:main-lower} only for $H \cap rS$. And to deal with $H \cap rS$, one proceeds
with the following steps for the pre-determined values of $m$ and $n$ that satisfy $N=mn$:
\begin{description}
\item{$(1)$} For the given choice of $0<\xi<1$, each individual function $h$
    satisfies a stable lower bound with parameters $(\xi/2,\ell,k)$.
\item{$(2)$} Given the $n$ blocks $(I_j)_{j=1}^n$ of cardinality $m$,
    with probability at least $1-2\exp(-c_0\eta n k)$, the stable lower bound in
    $(1)$ holds for at least $(1-\eta/2)n$ blocks.
\item{$(3)$} The high probability estimate in $(2)$ combined with the union bound
    allows one to obtain $(2)$ for a net in $H \cap rS$, as long as its cardinality
    is at most $\exp(c_1\eta nk)$ for $c_1=c_0/2$.
\item{$(4)$} If $\pi h$ denotes the nearest element to $h$ in the net,
    stability implies that for at least $(1-\eta/2)n$ of the blocks, one
    may discard the set $J_j(h)$ consisting of the $\ell$ largest values of the
    oscillation term $(|h-\pi h|(X_i))_{i \in I_j}$ and still have
    $$
    \frac{1}{m}\sum_{i \in I_j \backslash J_j(h)} (\pi h)^2(X_i) \geq
    \left(1-\frac{\xi}{2}\right)\|\pi h\|_{L_2}^2 =
    \left(1-\frac{\xi}{2}\right)r^2.
    $$
Hence, for every $h \in H \cap rS$ there are at least $(1-\eta/2)n$
blocks such that
\begin{align} \label{eq:the-argument}
& \Bigl(\frac{1}{m}\sum_{i=1}^m h^2(X_i)\Bigr)^{1/2}
\\
\geq & \Bigl(\frac{1}{m}\sum_{i
\in I_j \backslash J_j(h)} (\pi h)^2(X_i)\Bigr)^{1/2} - \Bigl(\frac{1}{m}\sum_{i \in
I_j \backslash J_j(h)} (h-\pi h)^2(X_i)\Bigr)^{1/2} \nonumber
\\
\geq & (1-\xi/2)^{1/2}r - \Bigl(\frac{1}{m}\sum_{i \in I_j \backslash J_j(h)} (h-\pi
h)^2(X_i)\Bigr)^{1/2} \geq (1-\xi)^{1/2}r \nonumber
\end{align}
where the last inequality holds if there is sufficient control on the last term.
\end{description}
Out of this list, $(1)$ is just the stable lower bound; $(2)$ is an immediate outcome
of Bennett's inequality; and $(3)$ is the reason for the entropy condition in Theorem
\ref{thm:main-lower}. This leaves us with the crucial point in the proof of Theorem
\ref{thm:main-lower}, which is establishing $(4)$.

To that end, let $\theta_1$ be a constant that is specified in what follows, and let
$H^\prime$ be a maximal $\theta_1 \xi r$-separated subset of $H \cap rS$.  Given a
sample $(X_i)_{i=1}^N$, let
$$
V = \left\{ v=\left((h - \pi h)(X_i) \right)_{i=1}^N : h \in H \cap rS \right\}
$$
and put $P_{I_j} v = (v_i)_{i \in I_j}$. The aim is to ensure that for every $v \in V$
there are at least $(1-\eta/2)n$ blocks $I_j$ such that
$$
\frac{1}{m}\sum_{i > \ell} \left((P_{I_j}v)^*_i\right)^2  \leq \frac{\xi^2}{4} r^2,
$$
implying that for every $h \in H \cap rS$, \eqref{eq:the-argument} holds for
$(1-\eta)n$ blocks.

In other words, if for $h \in H \cap rS$ and $v = \left((h - \pi h)(X_i)
\right)_{i=1}^N$ one sets
$$
\sharp_h =\left|\left\{j :  \frac{1}{m}\sum_{i > \ell} \left((P_{I_j}v)^*_i\right)^2
> \frac{\xi^2}{4} r^2\right\} \right|,
$$
then the main component of the proof of Theorem \ref{thm:main-lower} is to show that
with high probability,
$$
\sup_{h \in H \cap rS} \sharp_h \leq \frac{\eta n}{2}.
$$

\begin{Lemma} \label{lemma:single-to-uniform-osc}
There exist absolute constants $c_1,c_2$ and $c_3$ for which the following holds. Let
$H$ be star-shaped around $0$, set $\theta_1^2 \leq c_1\eta$  and let $H^\prime$ to be
a maximal $\theta_1 \xi r$-separated subset of $H \cap rS$ with respect to the
$L_2(\mu)$ norm. If
$$
\E \sup_{h \in H \cap rS} \left|\frac{1}{N} \sum_{i=1}^N \eps_i (h-\pi h)(X_i) \right|
\leq
c_2 \eta \sqrt{\frac{\ell}{m}} \xi r,
$$
then
$$
Pr \left(\sup_{h \in H \cap rS} \sharp_h > \frac{\eta n}{2}\right) \leq
2\exp\left(-c_3 N \frac{\ell}{m}
\min\left\{\eta,\frac{\eta^2}{\theta_1^2}\right\}\right).
$$
\end{Lemma}

The proof of Lemma \ref{lemma:single-to-uniform-osc} is based on Talagrand's
concentration inequality for empirical processes indexed by classes of uniformly
bounded functions \cite{MR1258865}, see also \cite{BoLuMa13}:
\begin{Theorem} \label{thm:Tal-conc}
There exists an absolute constant $C_0$ for which the following holds. Let $F$ be a
class of functions and set $\sigma^2_F = \sup_{f \in F} \E f^2$ and $b_F = \sup_{f \in
F} \|f\|_{L_\infty}$. Then, for any $x>0$, with probability at least $1-2\exp(-x)$,
\begin{equation} \label{eq:Tal:conc}
\sup_{f \in F} \left|\frac{1}{N}\sum_{i=1}^N f(X_i)-\E f \right| \leq C_0 \left(\E
\sup_{f \in F} \left|\frac{1}{N}\sum_{i=1}^N \eps_i f(X_i) \right| + \sigma_F
\sqrt{\frac{x}{N}} + b_F \frac{x}{N}\right).
\end{equation}
\end{Theorem}

\noindent {\bf Proof of Lemma \ref{lemma:single-to-uniform-osc}.}
Let $A=(m/\ell)^{1/2} \xi r$ and set
$$
\phi_A(t)=
\begin{cases}
A \cdot \sgn(t) &\mbox{if } \ \ |t| >A,
\\
t & \mbox{if } \ \ |t| \leq A.
\end{cases}
$$
Given $h \in H \cap rS$, $\pi h \in H^\prime$ and $v_i=(h-\pi h)(X_i)$ as above, let
$$
u_i=\phi_A(|v_i|) \ \ \ {\rm and} \ \ \ w_i =|v_i| \IND_{\{|v_i| > A \}} .
$$
Note that for every block $I_j$,
\begin{equation*}
\min_{J_j \subset I_j, |J_j|=\ell} \frac{1}{m} \sum_{i \in I_j \backslash J_j}
(P_{I_j} v)_i^2
\leq \min_{J_j \subset I_j, |J_j|=\ell} \frac{2}{m} \sum_{i \in I_j \backslash J_j}
u_i^2+w_i^2,
\end{equation*}
and if
$$
\frac{1}{m} \sum_{i \geq \ell} (P_{I_j} v_i^*)^2 =  \min_{J_j \subset I_j, |J_j|=\ell}
\frac{1}{m} \sum_{i \in I_j \backslash J_j} (P_{I_j} v)_i^2 \geq \xi^2 r^2/4
$$
then either
$\frac{1}{m} \sum_{i \in I_j} u_i^2 \geq \xi^2 r^2/16$, or, if the reverse inequality holds, there are at least $\ell$
coordinates in $I_j$ such that $w_i^2 \geq A^2$. Therefore, if we set
\begin{align*}
\sharp_h^1 = & \left|\left\{ j: \frac{1}{m} \sum_{i \in I_j} \phi_A^2 (|h-\pi h|(X_i))
\geq  \frac{\xi^2 r^2}{16} \right\} \right| \ \ \ {\rm and}
\\
\sharp_h^2 = & \left|\left\{j : (|h-\pi h|(X_i))_{\ell}^* \geq A
\right\}\right|,
\end{align*}
then
$$
\sharp_h \leq \sharp_h^1 + \sharp_h^2.
$$
Observe that if $\sup_{h \in H \cap rS} \sharp_h^1 \geq \eta n/4$ then
$$
\sup_{h \in H \cap r S} \sum_{j =1}^n \frac{1}{m}\sum_{i \in I_j} \phi_A^2 (|h-\pi
h|(X_i)) \geq \frac{\eta n}{4} \cdot \frac{\xi^2 r^2}{16};
$$
that is,
\begin{equation} \label{eq:cond-on-sharp-1}
(*)_1 \equiv \sup_{h \in H \cap r S} \frac{1}{N}\sum_{i=1}^N \phi_A^2 (|h-\pi h|(X_i))
\geq \frac{1}{64} \eta \xi^2 r^2.
\end{equation}
Invoking Theorem \ref{thm:Tal-conc}, let us show that with high probability, $(*)_1 <
\frac{1}{16} \eta \xi^2 r^2$, and therefore, on that event, $\sup_{h \in H \cap rS}
\sharp_h^1 < \eta n/4$.

Clearly, $\phi_A(t) \leq |t|$, and for $\theta_1^2 \leq c_1 \eta$ and for the right choice of $c_1$, it follows that
$$
\E  \phi_A^2 (|h-\pi h|(X_i)) \leq \E |h-\pi h|^2 \leq \theta_1^2 \xi^2 r^2 \leq
\frac{1}{256C_0} \eta \xi^2 r^2,
$$
where $C_0$ is the constant from \eqref{eq:Tal:conc}.

Also, $\phi_A^2$ is a Lipschitz function with a constant $2A$ and satisfies
$\phi_A^2(0)=0$. Thus, by the contraction inequality for Bernoulli processes
\cite{LeTa91}, one can ensure that
\begin{align*}
\E \sup_{h \in H \cap r S} \left| \frac{1}{N}\sum_{i=1}^N \eps_i \phi_A^2 (|h-\pi
h|(X_i)) \right| \leq & 2A \E \sup_{h \in H \cap r S} \left| \frac{1}{N}\sum_{i=1}^N
\eps_i (h-\pi h)(X_i) \right|
\\
\leq &\frac{1}{256C_0} \eta \xi^2 r^2
\end{align*}
provided that
\begin{equation} \label{eq:cond-2}
\E \sup_{h \in H \cap r S} \left| \frac{1}{N}\sum_{i=1}^N \eps_i (h-\pi h)(X_i)
\right| \lesssim \eta \xi^2 \frac{r^2}{A} \sim \left(\frac{\ell}{m}\right)^{1/2} \eta
\xi r
\end{equation}
by our choice of $A$.

Turning to the second term, note that for $F =\{\phi_A^2(|h-\pi h|(X)) : h \in H \cap r S\}$,
$$
\sigma_F^2 \leq \sup_{h \in H \cap r S} A^2 \|h-\pi h\|_{L_2}^2 \leq A^2 (\theta_1 \xi
r)^2 \ \ {\rm and} \ \ b_F \leq A^2.
$$
By Theorem \ref{thm:Tal-conc}, $(*)_1 \leq \frac{1}{16} \eta \xi^2 r^2$ with
probability at least
$$
1-2\exp\left(-c_3 N \frac{\xi^2 r^2}{A^2}
\min\left\{\eta,\frac{\eta^2}{\theta_1^2}\right\} \right),
$$
implying that
\begin{equation} \label{eq:sharp-1-est}
Pr \left(\sup_{h \in H \cap rS} \sharp_h^1 \leq \frac{\eta n}{4}\right) \geq
1-2\exp\left(-c N \frac{\ell}{m} \min\left\{\eta,\frac{\eta^2}{\theta_1^2}\right\}
\right).
\end{equation}
Next, note that if $\sharp_h^2 \geq \eta n/4$, then at least $\ell \eta n/4$ of the
values $(|h-\pi h|(X_i))_{i=1}^N$ are larger than $A$. To conclude the proof, let us
show that with high probability,
$$
\sup_{h \in H \cap rS} |\{i : |h-\pi h|(X_i) \geq A\}| \leq \frac{1}{8} \ell \eta n.
$$
Define $\Psi_A:\R_+ \to \R_+$ by
$$
\Psi_A(t) =
\begin{cases}
1 &\mbox{if } \ \ t \geq A,
\\
\frac{2}{A}\left(t - \frac{A}{2}\right) & \mbox{if } \ \ t \in [A/2,A),
\\
0 & \mbox{if } \ \ t \in [0,A/2].
\end{cases}
$$
It is evident that
$$
\sum_{i=1}^N \IND_{\{ |h-\pi h| \geq A\}}(X_i) \leq \sum_{i=1}^N \Psi_A (|h-\pi
h|(X_i)),
$$
and therefore, it suffices to show that
$$
\sup_{h \in H \cap rS} \frac{1}{N} \sum_{i=1}^N \Psi_A(|h-\pi h|(X_i)) \leq
\frac{1}{8} \eta \frac{\ell}{m}.
$$
Again, one may invoke Theorem \ref{thm:Tal-conc}. Observe that
$$
\E  \Psi_A (|h-\pi h|(X_i)) \leq Pr(|h-\pi h|(X) \geq A/2) \leq \frac{4\|h-\pi
h\|^2_{L_2}}{A^2} \leq \frac{4\theta_1^2 \xi^2 r^2}{A^2}.
$$
Therefore, $\E \Psi_A (|h-\pi h|(X_i)) \leq (\eta/24) \cdot (\ell/ m) $ provided that
\begin{equation} \label{eq:cond-3}
\frac{\theta_1^2 \xi^2 r^2}{A^2} \lesssim \eta \frac{\ell}{m},
\end{equation}
which, by our choice of $A$, holds if $\theta_1^2 \lesssim \eta$.

The function $\Psi_A(t)$ is Lipschitz with constant $2/A$ and $\Psi_A(0)=0$. By the
contraction inequality for Bernoulli processes,
\begin{align*}
\E \sup_{h \in H \cap rS} \left| \frac{1}{N}\sum_{i=1}^N \eps_i \Psi_A(|h-\pi h|(X_i))
\right| \leq & \frac{2}{A} \E \sup_{h \in H \cap rS} \left| \frac{1}{N}\sum_{i=1}^N
\eps_i (h-\pi h)(X_i) \right|
\\
\leq & \frac{\eta}{24C_0}\cdot \frac{\ell}{m}
\end{align*}
as long as
\begin{equation} \label{eq:cond-4}
\E \sup_{h \in H \cap rS} \left| \frac{1}{N} \sum_{i=1}^N \eps_i (h-\pi h)(X_i)
\right| \lesssim \eta \frac{\ell}{m} \cdot A \sim \eta \sqrt{\frac{\ell}{m}} \xi r,
\end{equation}
and \eqref{eq:cond-4} follows for our choice of $r$.

Moreover, for every $h \in H \cap rS$,
$$
\E \Psi_A^2 (|h-\pi h|(X)) \leq Pr(|h-\pi h|(X) \geq A/2) \leq \frac{4\theta_1^2 \xi^2
r^2}{A^2}
$$
and $\|\Psi_A( |h-\pi h|)\|_{L_\infty} \leq 1$. By Theorem \ref{thm:Tal-conc} and recalling once again that $A=(m/\ell)^{1/2} \xi r$, one has that
$$
Pr \left(\sup_{h \in H \cap rS} \frac{1}{N} \sum_{i=1}^N \Psi_A(|h-\pi h|(X_i)) \leq
\frac{1}{8}\eta \frac{\ell}{m} \right) \geq 1-2\exp\left(-c_6 N \frac{\ell}{m}
\min\left\{\eta,\frac{\eta^2}{\theta_1^2}\right\}\right),
$$
as claimed.
\endproof

Thanks to Lemma \ref{lemma:single-to-uniform-osc}, the proof of the first part of
Theorem \ref{thm:main-lower} follows by showing that there is a net $H^\prime$ of $H
\cap rS$ whose mesh width is $\theta_1 \xi r$, and with high probability, each $h^\prime \in
H^\prime$ satisfies an appropriate stable lower bound on at least $(1-\eta/2)n$ of the
blocks.

\begin{Lemma} \label{lemma:boosting-slb}
Let $k \geq \max\{4,2\log(4/\eta)\}$ and let $h$ satisfy a stable lower bound with
parameters $(\xi/2,\ell,k)$. Then with probability at least $1-2\exp(-\eta nk/8)$
there are at least $(1-\eta/2)n$ blocks $I_j$ such that for any $J_j
\subset I_j$ of cardinality $\ell$,
\begin{equation} \label{eq:in-lemma-boosting-slb}
\frac{1}{m}\sum_{i \in I_j \backslash J_j} h^2(X_i) \geq (1-\xi/2)\E h^2.
\end{equation}
Moreover, if $H^\prime$ is a class of functions that satisfy such a stable lower bound
and $\log |H^\prime| \leq \eta nk/16$, then with probability at least $1-2\exp(-\eta
nk/16)$, \eqref{eq:in-lemma-boosting-slb} holds for every $h^\prime \in H^\prime$.
\end{Lemma}

\proof Let $(\delta_j)_{j=1}^n$ be independent selectors that take the value $0$ on
the `good event' one is interested in: that is, each $\delta_j$ is a $\{0,1\}$-valued
random variable, defined by $\delta_j=0$ if for every $J_j \subset I_j$ of cardinality
at most $\ell$ one has
$$
\frac{1}{m}\sum_{i \in I_j \backslash J_j} h^2(X_i) \geq (1-\xi)\E h^2.
$$
Therefore, $\E \delta_j =\delta \leq 2\exp(-k)$. If $\eta \geq 4\exp(-k/2)$ and $k
\geq 4$ then by Bennett's inequality,
\begin{equation*}
Pr\left( \sum_{j=1}^n \delta_i \geq \frac{\eta}{2} n\right) \leq
\exp\left(-\frac{\eta}{2} n (\log(1+\eta/2\delta) -1)\right) \leq \exp(-\eta n k/8),
\end{equation*}
as required.

The second part of the claim is evident from the union bound.
\endproof

\vskip0.3cm

\noindent{\bf Proof of Theorem \ref{thm:main-lower}, part I.} As noted previously, the
claim is positive homogeneous, and since $H$ is star-shaped around $0$, it suffices to
prove it for $H \cap rS$. For that class, the combination of Lemma
\ref{lemma:boosting-slb} and Lemma \ref{lemma:single-to-uniform-osc} leads to the
wanted conclusion. Indeed, setting $\theta_1 \sim \sqrt{\eta}$, by Conditions $(1)$
and $(2)$ there is $H^\prime \subset H \cap rS$ that is $\theta_1 \xi r$-maximal
separated and $\log |H^\prime| \leq \eta nk/16$. Hence, by Lemma
\ref{lemma:boosting-slb}, with probability at least $$
1-2\exp(-\eta nk/4)=1-2\exp\left(-c_1 \eta N \frac{k}{m}\right),
$$
for every $u \in H^\prime$ there are at least $(1-\eta/2)n$ blocks $I_j$
such that for every $J_j \subset I_j$ of cardinality at most $\ell$,
\begin{equation} \label{eq:in-proof-main-1}
\frac{1}{m}\sum_{i \in I_j \backslash J_j} u^2(X_i) \geq (1-\xi/2) \|u\|_{L_2}^2 =
(1-\xi/2) r^2.
\end{equation}
Recall that $\pi h$ is the nearest point to $h$ in $H^\prime$ relative to the
$L_2(\mu)$ distance and set $v=\left((h-\pi h)(X_i)\right)_{i=1}^N$. By Lemma
\ref{lemma:single-to-uniform-osc}, with probability at least
$$
1-2\exp\left(-c_2 \eta N \frac{\ell}{m}\right),
$$
for every $h \in H \cap rS$, there are at most $\eta n/2$ blocks $I_j$ such
that
\begin{equation} \label{eq:in-proof-main-2}
\frac{1}{m} \sum_{i > \ell} \left((P_{I_j}v)_i^*\right)^2 > \frac{\xi^2 r^2}{2},
\end{equation}
and by \eqref{eq:the-argument}, if \eqref{eq:in-proof-main-1} and
\eqref{eq:in-proof-main-2} hold then for every $h \in H \cap rS$ there are at least
$(1-\eta)n$ blocks $I_j$ such that
$$
\frac{1}{m}\sum_{i \in I_j} h^2(X_i) \geq (1-\xi)r^2 = (1-\xi)\|h\|_{L_2}^2.
$$
\endproof

\vskip0.3cm
Let us turn to the proof of the second part of Theorem \ref{thm:main-lower}, showing
that Conditions $(2)$ and $(3)$ can be replaced by Conditions $(4)$ and $(5)$.

Clearly, if $r$ satisfies Condition $(4)$ for the right choice of constant then it
satisfies Condition $(3)$ as well. Therefore, all that is
left is to show that Conditions $(4)$ and $(5)$ also imply Condition $(2)$; in
particular, that if
\begin{equation*}
\E \sup_{u \in H \cap rD} \left|\frac{1}{N} \sum_{i=1}^N \eps_i u(X_i) \right| \leq
c\eta \sqrt{\xi} r \min\left\{\sqrt{\frac{\ell}{m}},\sqrt{\frac{k}{m}}\right\},
\end{equation*}
for the right choice of $c$ and under a stable lower bound, then
$$
\log{\cal M}(H \cap rS, c_0\sqrt{\eta} \xi r D) \leq \frac{\eta nk}{16},
$$
where, as always, ${\cal M}(F,\rho D)$ is the cardinality of a maximal $\rho$-separated subset of $F$ with respect to the $L_2(\mu)$ norm.

\begin{Theorem} \label{thm:entropy-to-Bernoulli}
There exist absolute constants $c_0,c_1$ and $c_2$ for which the following holds.
Let $\rho>0$ and $0<\eta<1$. Assume that for any $h_1,h_2 \in H \cap rS$ that are
$\rho$-separated, $h_1-h_2$ satisfies a stable lower bound with constants
$(1/2,\ell,k)$ for $k \geq c_0$. Assume further that
$$
\log {\cal M}(H \cap rS, \rho D) \geq \frac{\eta nk}{16}.
$$
Then with probability at least $1-2\exp(-c_1 \eta nk)$,
$$
\E_\eps \sup_{h \in H \cap rS} \left|\frac{1}{N}\sum_{i=1}^N \eps_i h(X_i) \right|
\geq c_2 \sqrt{\eta}\rho \min\left\{ \sqrt{\frac{\ell}{m}}, \sqrt{\frac{k}{m}}\right\}.
$$
\end{Theorem}
\begin{Remark}
The constant $1/2$ in the stable lower bound may be replaced by any number in $(0,1)$,
and that only affects the value of $c_2$ in Theorem \ref{thm:entropy-to-Bernoulli}.
\end{Remark}

Applying Theorem \ref{thm:entropy-to-Bernoulli} for the choice of $\rho = c_0
\sqrt{\eta}\xi r$ shows that Conditions $(4)$ and $(5)$ imply Condition $(2)$. With
that, the second part of the theorem follows from the first one.

\vskip0.4cm

The proof of Theorem \ref{thm:entropy-to-Bernoulli} is based on Sudakov's inequality
for Bernoulli processes \cite{LeTa91} in its scale-sensitive formulation (see, e.g.,
\cite{MR3274967}):
\begin{Theorem} \label{thm:Sudakov-for-Bernoulli}
There exists an absolute constant $c$ for which the following holds. Let $V \subset
\R^N$ and for every $v \in V$ set $Z_v = \sum_{i=1}^N \eps_i v_i$. If $|V| \geq
\exp(p)$ and $\{Z_v : v \in V\}$ is $\eps$-separated in $L_p$ then
$$
\E \sup_{v \in V} \sum_{i=1}^N \eps_i v_i \geq c \eps.
$$
\end{Theorem}

\noindent{\bf Proof of Theorem \ref{thm:entropy-to-Bernoulli}.} Let $h_1,h_2 \in H
\cap rS$ such that $\|h_1-h_2\|_{L_2} \geq \rho$, implying that $h_1-h_2$ satisfies a
stable lower bound with parameters $(1/2,\ell,k)$. Hence, by Lemma
\ref{lemma:boosting-slb}, with probability at least $1-2\exp(-c_1k n)$, there are at
least $n/2$ blocks $I_j$ such that for any $J_j \subset I_j$ of cardinality
$\ell$,
\begin{equation} \label{eq:Bernoulli-lower-1}
\frac{1}{m}\sum_{i \in I_j \backslash J_j} (h_1-h_2)^2(X_i) \geq \frac{1}{2} \E
(h_1-h_2)^2.
\end{equation}
Without loss of generality assume that the first $n/2$ blocks are among the
`good blocks', and that their union is $\{1,...,N/2\}$.
Set $v=(h_1(X_i))_{i=1}^N, \ u=(h_2(X_i))_{i=1}^N$, let $w=v-u$ and consider the random
variable
$$
Z_v-Z_u = \sum_{i=1}^N \eps_i w_i.
$$
By a standard contraction inequality \cite{LeTa91} and the characterization of the
$L_p$ norm of the random variable $Z_a=\sum_{i=1}^N \eps_i a_i$ from \cite{MR1244666},
it follows that for $p \geq 1$,
\begin{equation*}
\|Z_v-Z_u\|_{L_p} \gtrsim \| \sum_{i=1}^{N/2} \eps_i w_i \|_{L_p}
\sim 
\sum_{i \leq p} w_i^* + \sqrt{p} \left(\sum_{i=p+1}^{N/2}
\left(w_i^*\right)^2\right)^{1/2},
\end{equation*}
where $(w^*_i)_{i=1}^{N/2}$ denotes the nonincreasing rearrangement of $(|w_i|)_{i=1}^{N/2}$. Set $p=\ell n/4$ and let ${\cal I}$ be the set of indices of
the $p$ largest coordinates of $(|w_i|)_{i=1}^{N/2}$. Let $J_j = {\cal I} \cap I_j$ and observe that 
$$
\left| \left\{ j : |J_j| \leq \ell \right\} \right| \geq \frac{n}{4}.
$$   
Therefore, by \eqref{eq:Bernoulli-lower-1},
$$
\sum_{i=p+1 }^{N/2}  \left(w_i^*\right)^2 \geq \frac{n}{4} \cdot
\frac{m}{2} \|h_1-h_2\|_{L_2}^2 \gtrsim N \|h_1-h_2\|_{L_2}^2.
$$
It follows that for every sample $(X_i)_{i=1}^N$ in an event with
probability at least $1-2\exp(-c_1kn)$,
\begin{equation} \label{eq:other-p}
\|Z_v-Z_u\|_{L_p} \gtrsim \sqrt{p} \cdot \sqrt{N} \|h_1-h_2\|_{L_2},
\end{equation}
and clearly, on the same event, \eqref{eq:other-p} holds for any $1 \leq p \leq \ell n/4$. 

Now, let $H^\prime$ be a maximal $\rho$-separated subset of $H \cap rS$ and recall
that $\log |H^\prime| \gtrsim \eta nk$. By the union bound, with probability at least
$1-2\exp(-c_2\eta nk)$ the random set $V=\{ (h(X_i))_{i=1}^N : h \in H^\prime\}$ contains at
least $\exp(c_3\eta nk)$ vectors $v$, for which the random variables $\sum_{i=1}^N
\eps_i v_i$ are $\sim \sqrt{p} \cdot \sqrt{N} \rho $ separated in $L_p$ for any $2
\leq p \leq \ell n/4$. Set
$$
p \sim n \min\{\eta k,\ell\} = N \min \left\{\frac{\eta k}{m},\frac{\ell}{m}\right\}
$$
and by Theorem \ref{thm:Sudakov-for-Bernoulli}, with probability at least
$1-2\exp(-c_4 \eta nk)$ relative to $X_1,...,X_N$,
$$
\E_\eps \sup_{v \in V} Z_v \geq c_5 N \rho \min \left\{\sqrt{\frac{\ell}{m}},
\sqrt{\frac{\eta k}{m}}\right\},
$$
implying that
$$
\E_\eps \sup_{h \in H \cap rS} \left|\frac{1}{N} \sum_{i=1}^N \eps_i h(X_i) \right|
\geq c_5 \sqrt{\eta} \rho \min \left\{\sqrt{\frac{\ell}{m}},\sqrt{\frac{k}{m}}\right\}.
$$
\endproof

\bibliographystyle{plain}
\bibliography{SBM}

\end{document}